\documentclass[10pt,twocolumn]{article}
\usepackage[letterpaper,margin=0.72in,columnsep=0.24in]{geometry}
\usepackage{amsmath,amssymb,graphicx,booktabs,array,multirow}
\usepackage[numbers,sort&compress]{natbib}
\usepackage{xcolor,microtype,caption}
\usepackage[colorlinks=true,citecolor=blue!55!black,linkcolor=blue!55!black,urlcolor=blue!55!black]{hyperref}
\definecolor{methodblue}{HTML}{2C6095}
\newcommand{\method}{CARE-Drive}
\hypersetup{pdftitle={CARE-Drive: Component-wise Alternating Refinement of Ego-Trajectories for Vision-Language-Action Driving},pdfauthor={Xiao Ma, Yijun Chen, Zhiquan Hu, Jicheng Zhao, Ran Ji, Yuming Li, Chuang Dai}}
\newcommand{\R}{\mathbb{R}}
\newcommand{\ind}{\mathbb{I}}
\newcommand{\sg}{\operatorname{sg}}
\newcommand{\wrap}{\operatorname{wrap}}

\title{\vspace{-1.2em}\textbf{CARE-Drive: Component-wise Alternating Refinement\\of Ego-Trajectories for Vision-Language-Action Driving}}
\author{Xiao Ma\thanks{These authors contributed equally.}, Yijun Chen\footnotemark[1], Zhiquan Hu,\\
Jicheng Zhao, Ran Ji, Yuming Li\thanks{Corresponding author.}, Chuang Dai\\[0.35em]
\normalsize NEXTAI Research Institute, Chery Group}
\date{October 27, 2026}

\begin{document}
\maketitle

\begin{abstract}
Vision--language--action (VLA) models bring multimodal understanding and world knowledge to end-to-end autonomous driving. To further improve the quality of predicted trajectories, existing studies introduce candidate refinement, residual denoising, and token correction. However, these methods typically focus on unified refinement of the entire trajectory, paying limited attention to the distinct physical quantities and their interactions. We further investigate physical specialization and coordination in trajectory refinement, centering on position and heading for simpler and more intuitive correction modeling. To this end, we propose \method{} (Component-wise Alternating Refinement of Ego-Trajectories), which decomposes trajectory refinement into waypoint residual correction and position-conditioned heading generation. Through imitation learning and a future-token auxiliary task, the backbone provides predictive representations and a base trajectory. The position expert predicts bounded metric waypoint residuals, while the heading expert reconstructs vehicle orientation from adjacent angular increments conditioned on the corrected path. With the backbone frozen, the two experts are optimized alternately through supervised initialization and reinforcement fine-tuning. This design simplifies refinement through clear physical specialization while preserving conditional coupling during prediction and optimization. On NAVSIM v1, \method{} achieves 88.8 PDMS, improving by 2.4 points over the unrefined base trajectories generated by the same backbone. Ablation studies further support the effectiveness of specialized experts and staged training. The source code is available at \url{https://github.com/marshal12138/CARE-Drive}.
\end{abstract}

\section{Introduction}
Autonomous driving models draw on complementary perception and prediction architectures. Convolutional and Transformer-based encoders extract camera or bird's-eye-view (BEV) features for planning~\cite{uniad}, while vision--language models (VLMs) provide pretrained multimodal understanding and a language-generation interface~\cite{autovla,qwen3vl}. World models capture temporal structure through video generation~\cite{drivelaw,epona} or latent prediction. Joint-embedding predictive architecture (JEPA) methods predict target representations without requiring pixel reconstruction~\cite{vjepa}. CoWorld-VLA~\cite{coworld} illustrates how predictive representations can complement a VLM backbone.

VLMs are particularly attractive because their language interface supports instruction following and visual question answering. Future-feature supervision can additionally train the same backbone to anticipate scene evolution in latent space~\cite{coworld}. This provides an architectural basis for combining world-future reasoning with language interaction, while exposing shared features for action planning.

Action generation and refinement can use direct trajectory regression, autoregressive physical-action decoding~\cite{autovla}, iterative diffusion~\cite{diffusiondrive}, or candidate evaluation~\cite{wote}. Reinforcement fine-tuning further adapts policies to planning rewards~\cite{autovla}. Residual correction offers a complementary approach: adjusting an initial trajectory with targeted updates. For physically interpretable refinement, the key question is how to assign these updates to quantities with distinct roles, such as position and heading.

As shown in Figure~\ref{fig:framework}, we propose \method{}, a future-supervised VLA framework with physically meaningful correction experts. A Qwen3-VL-2B backbone learns predictive representations and an initial trajectory. A position expert predicts bounded XY residuals, and a heading expert reconstructs orientation from adjacent angular increments conditioned on corrected positions and shared features. This gives each expert an explicit physical target and makes heading refinement responsive to position changes. The experts are optimized with the backbone frozen, keeping correction learning modular and the backbone's language-generation interface structurally intact. Deployment produces one trajectory through greedy decoding.

Our contributions are twofold:
\begin{itemize}
\item \textbf{An effective world--action framework.} We propose \method{}, integrating future-supervised representation learning with staged trajectory refinement for end-to-end driving. Experiments on NAVSIM v1 demonstrate favorable planning performance.
\item \textbf{Physically meaningful correction experts.} We design complementary position and heading experts that output metric waypoint residuals and adjacent angular increments. Corrected positions condition heading generation, giving spatial and angular refinement explicit physical meaning.
\end{itemize}

\section{Related Work}
\paragraph{End-to-end and generative planning.}
UniAD~\cite{uniad} studies planning-oriented perception and prediction. Hydra-MDP~\cite{hydra} and DiffusionDrive~\cite{diffusiondrive} introduce multimodal planning and generative trajectory modeling. \method{} uses a continuous trajectory anchor followed by categorical refinement. Section~\ref{sec:decoder-sharing} compares its end-to-end latency with a parameter-matched single correction expert.

\paragraph{World models for driving.}
LAW~\cite{law} learns latent world representations for planning. Epona~\cite{epona}, PWM~\cite{pwm}, WoTE~\cite{wote}, and DriveLaW~\cite{drivelaw} investigate generative, predictive, or evaluative links between future states and actions. CoWorld-VLA~\cite{coworld} transfers complementary semantic, geometric, dynamic, and trajectory information into expert tokens, and combines them with a hierarchical fusion planner. Its framework is the closest reference for organizing our external comparisons. Our implementation draws on the CoWorld codebase and uses three camera histories, one future-feature teacher, an MLP anchor, and two categorical action experts. Here, \emph{world--action modeling} couples predictive latent supervision to action learning.

\paragraph{VLA and reinforcement fine-tuning.}
AutoVLA~\cite{autovla} uses autoregressive physical actions and adaptive language reasoning. ReCogDrive~\cite{recogdrive}, DriveVLA-W0~\cite{drivevla}, and Uni-World VLA~\cite{uniworld} provide further examples of cognitive or world-informed planning. Our action experts use group-relative policy optimization inspired by GRPO~\cite{grpo} and a clipped token-level surrogate related to PPO~\cite{ppo}. The heading-stage greedy baseline is inspired by self-critical sequence training~\cite{scst}. We combine these objectives with batch-level RMS calibration of candidate advantages.

\section{Method}
\subsection{Task and Notation}
For scene $i$, the observation is $o_i=(I_i,s_i,r_i)$, where $I_i$ denotes camera history, $s_i$ the ego-state description, and $r_i$ the high-level route command. The history contains four frames from each of three cameras (front, front-left, and front-right), sampled at $-1.5,-1.0,-0.5,0$ seconds at 2 Hz. The task is to predict an ego trajectory from this observation:
\begin{equation}
\begin{aligned}
\widehat T_i&=f_{\Theta}(o_i),\\
T_i^*&=\{(x_{it}^*,y_{it}^*,\psi_{it}^*)\}_{t=1}^{N_w}.
\end{aligned}
\end{equation}
Here $\Theta$ comprises the backbone and correction experts, and $T_i^*$ is the ground-truth target. Each trajectory contains $N_w=8$ positions and headings at 0.5-second intervals over four seconds, expressed in the local ego frame. Position is measured in meters and heading in radians. Subscripts $i,t$ index scenes and future steps; superscript $\mathrm{base}$ denotes the backbone prediction, $*$ denotes ground truth, and a hat denotes a model prediction. $X_i^{\mathrm{base}}$ and $\widehat X_i$ contain the corresponding XY positions.

During backbone training, \method{} couples trajectory imitation with a future-token auxiliary task. Given $o_i$, future query tokens learn to predict latent features of future images, with targets supplied by a frozen video teacher. Action query tokens attend to the future-query states through causal attention and provide the features used to generate $T_i^{\mathrm{base}}$. Joint optimization of future-feature prediction and trajectory regression trains the shared backbone to encode scene evolution for planning. The trained backbone then supplies the base trajectory and shared action features to the coupled position and heading experts.

\begin{figure*}[t]
\centering
\includegraphics[width=\textwidth]{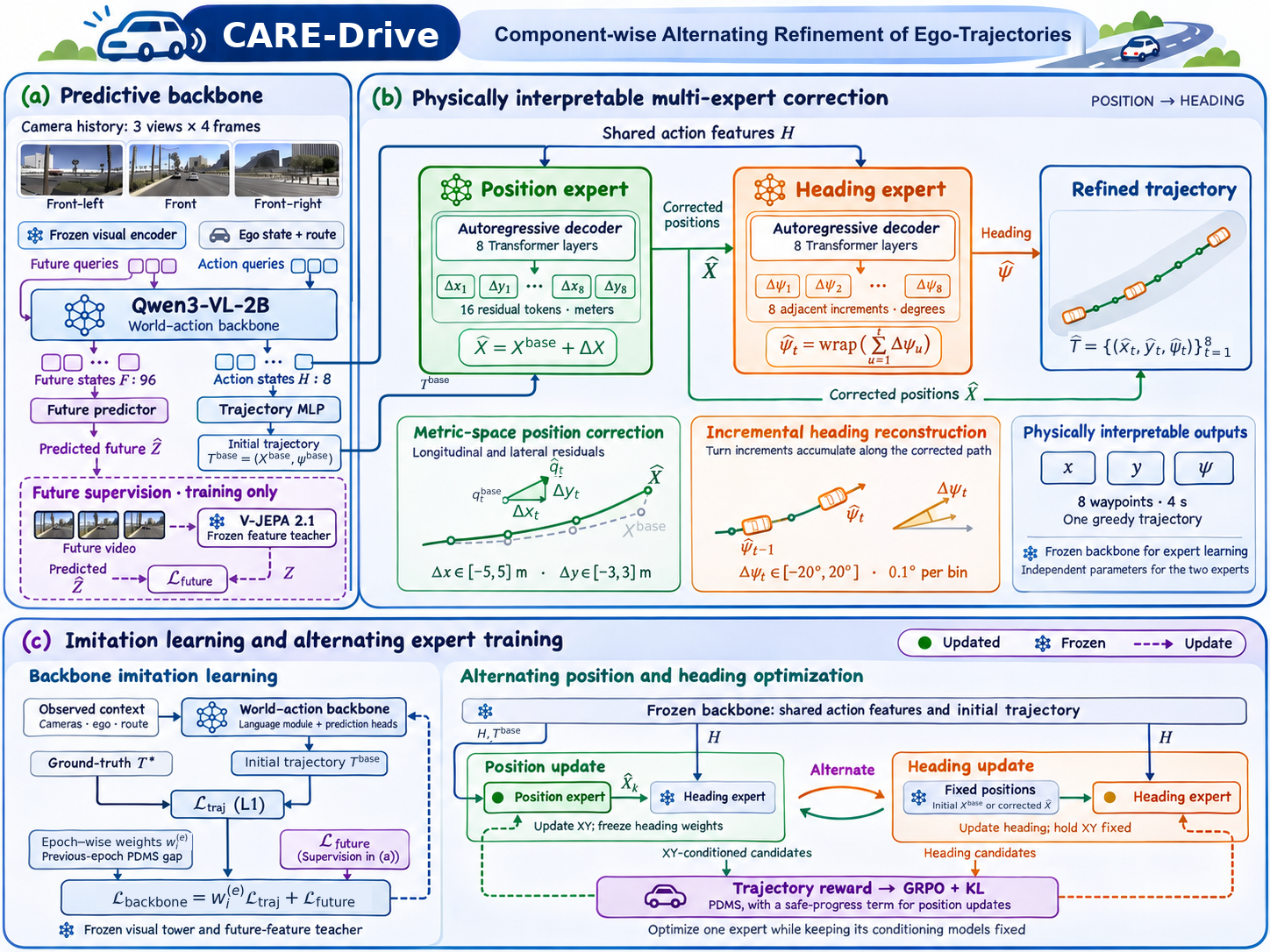}
\caption{\textbf{\method{} overview.} (a) The predictive backbone produces shared action features and an initial trajectory; a frozen video teacher provides future-feature supervision during training. (b) The position expert predicts metric waypoint residuals, and the heading expert reconstructs orientation from adjacent angular increments conditioned on corrected positions and shared action features. Greedy decoding yields one eight-waypoint trajectory. (c) Dynamically weighted imitation learning trains the backbone, followed by alternating position and heading optimization with the backbone frozen. Each expert update holds its conditioning models fixed. Snowflakes mark frozen modules.}
\label{fig:framework}
\end{figure*}

\subsection{Future-Token Auxiliary Task}
The backbone uses a frozen visual tower and a trainable language module. Learnable future query tokens $Q_f\in\R^{N_f\times d}$ and action query tokens $Q_a\in\R^{N_a\times d}$ follow the observation tokens in the causal input sequence. We use $N_f=96$, $N_a=8$, and hidden dimension $d=2048$. The future queries use four temporal token identities, each repeated at 24 spatial positions; the eight action queries have distinct token identities. Their embeddings are learned with the language backbone. Writing $\theta$ for the backbone parameters, the corresponding contextual hidden states are
\[
\begin{aligned}
(F_i,H_i)&=B_{\theta}(o_i;Q_f,Q_a),\\
F_i&\in\R^{N_f\times d},\quad H_i\in\R^{N_a\times d}.
\end{aligned}
\]
As shown in Figure~\ref{fig:framework}(a), a two-layer projection predicts teacher features $\widehat Z_i=P(F_i)\in\R^{N_f\times d_z}$, with $d_z=1664$. The base MLP reads the last action state $H_i[N_a]$ and produces $T_i^{\mathrm{base}}=G(H_i[N_a])$. Future-query information reaches the action states through the backbone's causal attention.

A frozen V-JEPA 2.1 ViT-G encoder supplies training targets $Z_i\in\R^{N_f\times d_z}$, following the predictive-feature direction of V-JEPA~\cite{vjepa}. It encodes eight future images resized to $384\times384$. The resulting feature grid has four temporal groups, each spatially pooled to $4\times6$, giving 96 vectors of dimension 1,664. The auxiliary future-feature prediction loss is
\begin{align}
\mathcal L_{\mathrm{future},i}
&=\mathrm{SmoothL1}(\widehat Z_i,\sg(Z_i))\nonumber\\
&\quad+\frac{0.5}{N_f}\sum_{j=1}^{N_f}\left[1-\cos(\widehat Z_{ij},\sg(Z_{ij}))\right].
\label{eq:future}
\end{align}
SmoothL1 uses a transition value of one and averages over tokens and features. The teacher and future images provide training supervision; inference uses the observed camera history, ego state, and route command.

\subsection{Epoch-Wise Difficulty-Weighted Imitation}
The base trajectory loss averages the absolute errors of the two position coordinates and heading over all predicted waypoints:
\begin{equation}
\begin{aligned}
\ell_i^{\mathrm{traj}}&=\frac{1}{3N_w}\sum_{t=1}^{N_w}
\Bigl(|x_{it}^{\mathrm{base}}-x_{it}^*|+|y_{it}^{\mathrm{base}}-y_{it}^*|\\
&\qquad\qquad+|\psi_{it}^{\mathrm{base}}-\psi_{it}^*|\Bigr).
\end{aligned}
\end{equation}
Position errors are computed in meters and heading errors in radians. Each component enters directly in these physical units with coefficient one.

The first epoch uses uniform weights. Before epochs two through five, the previous frozen epoch endpoint predicts trajectories on the scoreable training population $\mathcal S$. Let $R_i^*$ and $R_i^{(e-1)}$ be official normalized PDMS rewards for the ground-truth trajectory and the model trajectory. Define
\begin{align}
h_i^{(e)}&=\ind\!\left[R_i^*-R_i^{(e-1)}>0.01\right],\\
\widetilde w_i^{(e)}&=1+h_i^{(e)},\qquad
w_i^{(e)}=\frac{\widetilde w_i^{(e)}}{|\mathcal S|^{-1}\sum_{j\in\mathcal S}\widetilde w_j^{(e)}}.
\label{eq:weights}
\end{align}
Weights remain fixed within epoch $e$. For $N$ training scenes, the objective is
\begin{equation}
\mathcal L_{\mathrm{backbone}}=
\frac{1}{N}\sum_{i=1}^{N}\left(w_i^{(e)}\ell_i^{\mathrm{traj}}+\mathcal L_{\mathrm{future},i}\right).
\end{equation}
The weights apply to the trajectory term; the future-feature loss retains coefficient one.

\subsection{Position and Heading Experts}
After backbone training, $H_i$ and $T_i^{\mathrm{base}}$ are fixed. The position expert is an autoregressive Transformer conditioned on both. It generates 16 tokens in interleaved $(x_1,y_1,\ldots,x_8,y_8)$ order. Each axis has 401 categories, with index $k_{it,a}\in\{0,\ldots,400\}$ for coordinate $a\in\{x,y\}$. Let $q_{it,x}^{\mathrm{base}}=x_{it}^{\mathrm{base}}$ and $q_{it,y}^{\mathrm{base}}=y_{it}^{\mathrm{base}}$. Training-data residual and quantization analyses support the adopted bounds $b_x=5$ m and $b_y=3$ m; the analysis is detailed in Appendix~\ref{app:interfaces}. Residual decoding is
\begin{equation}
\begin{aligned}
\Delta q_{it,a}&=b_a\frac{k_{it,a}-200}{200},\\
\widehat q_{it,a}&=q_{it,a}^{\mathrm{base}}+\Delta q_{it,a},\quad a\in\{x,y\}.
\end{aligned}
\end{equation}
We use fixed symmetric residual bounds with 401 bins per axis, yielding resolutions of 0.025 m in X and 0.015 m in Y.

Supervised labels are formed by clipping and rounding the target residuals. The expert also receives normalized anchor displacement, heading, and time features, as detailed in Appendix~\ref{app:interfaces}.

The heading expert receives $H_i$ and the corrected sequence $\widehat X_i\in\R^{N_w\times2}$ from the position expert (Figure~\ref{fig:framework}(b)). Analysis of adjacent heading increments in training samples supports the adopted $[-20^{\circ},20^{\circ}]$ range (Appendix~\ref{app:interfaces}). The expert generates eight increments, each uniformly quantized in $0.1^{\circ}$ steps into 401 values with indices $k_{it,\psi}\in\{0,\ldots,400\}$:
\begin{equation}
\Delta\psi_{it}=\frac{\pi}{1800}(k_{it,\psi}-200),\qquad
\widehat\psi_{it}=\wrap\!\left(\sum_{u=1}^{t}\Delta\psi_{iu}\right).
\label{eq:heading}
\end{equation}
The heading sequence is reconstructed by accumulating predicted increments from zero in the local ego frame, conditioned on shared action features and corrected XY positions. Each expert has eight layers, width 768, eight attention heads, and an FFN width of 3,072. The position and heading experts each contain approximately 78.42M parameters, paired with a Qwen3-VL-2B backbone.

During position-policy training, the frozen preliminary heading expert generates a matching heading sequence for each candidate XY trajectory, and the reward is evaluated on the resulting complete trajectory. Subsequent heading adaptation starts from that preliminary expert and holds the final position expert and its greedy XY fixed.

\subsection{Rewards and Expert Optimization}
\label{sec:reward-optimization}
\paragraph{Training reward.}
Let $R$ denote the normalized NAVSIM v1 planning reward. Position refinement uses
\begin{equation}
R^{\mathrm{pos}}=R+\lambda\ind[\mathrm{NC}=\mathrm{DA}=\mathrm{TTC}=\mathrm{C}=1]\,
\mathrm{EP}(2-\mathrm{EP}),
\label{eq:safeep}
\end{equation}
where NC, DA, TTC, and C assess collision, road, time-to-collision, and comfort compliance, and EP measures progress. We use $\lambda=0$ in the first position RL stage and $\lambda=1$ in the second, encouraging progress only among compliant candidates. Evaluation always uses the original reward $R$; coefficient comparisons are reported in Section~\ref{sec:progress-reward-ablation}.

\paragraph{Candidate advantages.}
For each scene $i$, we sample eight trajectories. Let $v_{ik}=(R_{ik}-\overline R_i)/\sigma_i$ be the standard PDMS-based advantage, with $\sigma_i$ the numerically stabilized within-scene reward standard deviation. Preliminary heading training uses $A_{ik}=v_{ik}$; position training and final heading adaptation use
\begin{equation}
u_{ik}=
\begin{cases}
R^{\mathrm{pos}}_{ik}-\overline R_i^{\mathrm{pos}}, & \text{position},\\[2pt]
(R_{ik}-R_i^{\mathrm{greedy}})/\sigma_i, & \text{heading adaptation},
\end{cases}
\label{eq:expert-advantage}
\end{equation}
where $R_i^{\mathrm{greedy}}$ is the current policy's greedy reward, treated as a constant. Thus, position candidates are compared with their group mean, while final heading candidates are compared with the policy's greedy output.

\paragraph{Scale calibration and policy update.}
These two stages apply a shared batch-wise scale,
\begin{equation}
A_{ik}=u_{ik}\,
\frac{\operatorname{RMS}_{\mathcal B}(v)}
     {\operatorname{RMS}_{\mathcal B}(u)},
\label{eq:rms}
\end{equation}
where $\mathcal B$ contains the batch's candidate trajectories. The scaling preserves advantage signs and relative magnitudes while keeping the overall learning-signal scale comparable across reward settings. Experts are updated using a clipped GRPO objective with KL regularization toward the stage's fixed reference policy~\cite{grpo,ppo}.

\subsection{Training Curriculum and Inference}
The backbone imitation and alternating expert optimization in Figure~\ref{fig:framework}(c) comprise eight five-epoch stages: backbone imitation; heading SFT followed by two heading RL stages; position SFT followed by two position RL stages; and final heading adaptation.

Both preliminary heading RL stages optimize PDMS on the backbone's fixed XY. The first starts from the heading SFT endpoint; the second starts from the first RL endpoint and uses it as a refreshed, fixed KL reference. The reward and advantage definition remain unchanged, while the optimizer and scheduler are restarted for the new stage.

Position training keeps this preliminary heading expert frozen. The first position RL stage optimizes PDMS; the second introduces the safety-gated progress bonus. Finally, with the trained position expert and its greedy XY fixed, the heading expert is adapted using the greedy-baseline, RMS-calibrated update described above. Backbone training has 3,625 updates; each expert stage has 3,620 updates, for a total of 28,965 optimizer updates.

Inference computes $H_i,T_i^{\mathrm{base}}$, greedily decodes 16 position tokens, then greedily decodes eight heading increments.

\section{Experimental Results}
We evaluate on NAVSIM v1~\cite{navsim}, a trajectory-planning benchmark built from real-world driving logs. It provides sensor observations and ego states for predicting future ego trajectories. Evaluation uses non-reactive simulation, in which other traffic participants follow their recorded motion. Planning performance is measured by the PDM Score (PDMS), which combines safety, driving progress, and comfort. We report the mean scores for no-at-fault collisions (NC), drivable-area compliance (DAC), time-to-collision compliance (TTC), comfort (Comf.), and ego progress (EP), alongside PDMS; all are expressed on a 0--100 scale, with higher values better. PDMS is averaged from per-scene composite scores.

\subsection{Test-Set Performance}
\begin{table*}[t]
\centering\small
\caption{\textbf{NAVSIM v1 navtest results.} AutoVLA and \method{} are locally evaluated on 12,146 scenes using the official CPU scorer; other results follow CoWorld-VLA~\cite{coworld}. C = camera, L = LiDAR; $N$ denotes multiple frames when the exact count is not specified by the source. All scores use a 0--100 scale, where higher is better. Values are displayed to one decimal; trailing zeros for reported integers are formatting only.}
\label{tab:literature}
\setlength{\tabcolsep}{3pt}
\begin{tabular}{llccrrrrrr}
\toprule
Family & Method & Sensors & Frames & NC $\uparrow$ & DAC $\uparrow$ & TTC $\uparrow$ & Comf. $\uparrow$ & EP $\uparrow$ & PDMS $\uparrow$\\
\midrule
End-to-end & UniAD~\cite{uniad} & C & $N$ & 97.8 & 91.9 & 92.9 & 100.0 & 78.8 & 83.4\\
 & Hydra-MDP~\cite{hydra} & C+L & $N$ & 98.3 & 96.0 & 94.6 & 100.0 & 78.7 & 86.5\\
 & DiffusionDrive~\cite{diffusiondrive} & C+L & $N$ & 98.2 & 96.2 & 94.7 & 100.0 & 82.2 & 88.1\\
\midrule
World models & LAW~\cite{law} & C & $N$ & 96.4 & 95.4 & 88.7 & 99.9 & 81.7 & 84.6\\
 & Epona~\cite{epona} & C & $N$ & 97.9 & 95.1 & 93.8 & 99.9 & 80.4 & 86.2\\
 & PWM~\cite{pwm} & C & $N$ & 98.6 & 95.9 & 95.4 & 100.0 & 81.8 & 88.1\\
 & WoTE~\cite{wote} & C+L & $N$ & 98.5 & 96.8 & 94.9 & 99.9 & 81.9 & 88.3\\
 & ResWorld~\cite{resworld} & C+L & $N$ & 98.9 & 96.5 & 95.6 & 100.0 & 83.1 & 89.0\\
 & DriveLaW~\cite{drivelaw} & C & $N$ & 99.0 & 97.1 & 96.7 & 100.0 & 81.3 & 89.1\\
\midrule
VLM / VLA & ReCogDrive (w/o RL)~\cite{recogdrive} & C & 1 & 98.1 & 94.7 & 94.2 & 100.0 & 80.9 & 86.5\\
 & DriveVLA-W0~\cite{drivevla} & C & $N$ & 98.4 & 95.3 & 95.4 & 100.0 & 80.9 & 87.2\\
 & AutoVLA~\cite{autovla} & C & 4 & 99.0 & 94.8 & 96.9 & 99.8 & 81.4 & 88.0\\
 & Uni-World VLA~\cite{uniworld} & C & $N$ & 98.7 & 96.7 & 96.1 & 100.0 & 83.2 & 89.4\\
 & CoWorld-VLA~\cite{coworld} & C & 1 & 99.1 & 97.0 & 96.5 & 100.0 & 84.0 & 90.0\\
\midrule
This work & \textbf{\method{}} & C & 4 & 98.4 & 95.9 & 95.3 & 100.0 & 84.0 & 88.8\\
\bottomrule
\end{tabular}
\end{table*}

\method{} achieves 88.837 PDMS, placing third among the VLM/VLA methods listed in Table~\ref{tab:literature}. Using unrounded scores, its locally evaluated PDMS is 0.811 points above AutoVLA's 88.026 and approximately 1.2 points below CoWorld-VLA's reported 90.0. Compared with local AutoVLA, \method{} scores higher on DAC, comfort, and EP, and lower on NC and TTC. The paired log-bootstrap 95\% interval for the PDMS difference from AutoVLA is $[-0.148,1.734]$ and includes zero; it describes uncertainty for the fixed evaluated models. A complete backbone-and-expert training workflow takes approximately 19 hours on 16 Ascend 910 accelerators.

\subsection{Effect of Expert Refinement}
\label{sec:expert-ablation-overview}
Table~\ref{tab:expert-components} summarizes the planning gains from progressively adding correction experts to the same frozen backbone. The experts are trained on 92,614 scenes, and all four configurations are evaluated on the complete 12,146-scene \texttt{navtest} split. The complete refinement pipeline increases PDMS from 86.462 to 88.837, a gain of 2.375 points over the unrefined backbone trajectory.

\begin{table*}[htbp]
\centering\small
\caption{Effect of progressively adding correction experts on the full test set. Gains are relative to the backbone-only configuration.}
\label{tab:expert-components}
\setlength{\tabcolsep}{7pt}
\begin{tabular}{lrr}
\toprule
Configuration & Full-test PDMS $\uparrow$ & Gain over backbone\\
\midrule
Backbone only & 86.462 & ---\\
Add heading expert & 87.562 & $+1.100$\\
Add position expert; freeze heading expert & 88.672 & $+2.210$\\
Freeze position expert; adapt heading expert & \textbf{88.837} & $+2.375$\\
\bottomrule
\end{tabular}
\end{table*}

The heading expert first preserves the backbone's XY waypoints and changes only their headings, raising PDMS to 87.562, an increase of 1.100 points. The position expert then undergoes supervised initialization and reward-based training with the preliminary heading expert frozen; each corrected path receives its corresponding conditional headings. This stage raises PDMS to 88.672, a further gain of 1.110 points. Finally, with the corrected XY fixed, the heading expert adapts to its new conditioning inputs and reaches 88.837, adding 0.165 points.

\subsection{Ablation Studies}
\label{sec:ablations}

\subsubsection{Effect of Future Queries and Supervision}
\label{sec:future-ablation}
We examine future queries and future-feature supervision using 8,933 sampled training scenes and an 885-scene sampled test set, with an L1 trajectory loss. The three variants remove the future branch, retain 96 future queries without future supervision, or retain the same queries with the future-token auxiliary task. The three variants share the sampled data and the remaining training settings, and are evaluated before expert correction.

\begin{table}[htbp]
\centering\small
\caption{Effect of the future-token auxiliary task.}
\label{tab:future-ablation}
\setlength{\tabcolsep}{4pt}
\begin{tabular}{lccr}
\toprule
Variant & Future queries & Supervision & PDMS $\uparrow$\\
\midrule
No future branch & 0 & No & 74.310\\
Queries only & 96 & No & 74.680\\
Auxiliary task & 96 & Yes & \textbf{76.800}\\
\bottomrule
\end{tabular}
\end{table}

Table~\ref{tab:future-ablation} shows a 0.370-point increase from adding future queries alone. With the query structure fixed, future supervision adds a further 2.120 points, bringing the total improvement over the variant without a future branch to 2.490 points. This ablation demonstrates the value of the future-token auxiliary task for trajectory planning.

\subsubsection{Visual-Tower Unfreezing}
\label{sec:visual-unfreeze}
We assess visual-tower unfreezing on 8,933 sampled training scenes and the 885-scene sampled test set. This comparison retains the original trajectory-MSE training setup. The frozen and unfrozen variants start from the same pretrained backbone and prediction-head initialization, with identical $1024\times512$ video inputs, future supervision, and a ten-epoch budget of 700 updates, and the same training seed. The unfrozen variant also trains the visual merger, while the future-feature teacher remains frozen in both variants. We evaluate the final-epoch backbone trajectories before expert correction.

\begin{table}[htbp]
\centering\small
\caption{Effect of visual-tower freezing.}
\label{tab:visual-unfreeze}
\begin{tabular}{lr}
\toprule
Student vision tower & Test PDMS $\uparrow$\\
\midrule
Frozen & \textbf{74.198}\\
Unfrozen (including merger) & 69.097\\
\bottomrule
\end{tabular}
\end{table}

Unfreezing reduces test PDMS by 5.101 points (Table~\ref{tab:visual-unfreeze}), with a paired log-bootstrap 95\% interval of $[-7.646,-2.548]$. This comparison supports keeping the vision tower frozen under the evaluated training configuration.

\subsubsection{Difficulty-Weighted Imitation Learning}
\label{sec:weight-ablation}
To evaluate difficulty weighting with a larger training set, we train on all 92,772 eligible scenes in our NAVSIM v1 training split and evaluate on the complete \texttt{navtest} split of 12,146 scenes. We compare uniform weighting with dynamic difficulty weighting under the same pretrained initialization, training seed, five-epoch budget, trajectory L1 objective, and future supervision. Uniform weighting assigns unit weights to all scenes. Dynamic weighting starts with uniform weights and then applies the epoch-wise PDMS-gap rule in Equation~\ref{eq:weights}, emphasizing scenes where the previous endpoint's trajectory scores more than one PDMS point below the ground truth. Evaluation uses the fifth-epoch backbone trajectories before expert correction.

\begin{table}[htbp]
\centering\small
\caption{Difficulty-weighted imitation learning. Full-test PDMS of the backbone without correction experts.}
\label{tab:weight-ablation}
\setlength{\tabcolsep}{4pt}
\begin{tabular}{lr}
\toprule
Training & Full-test PDMS\\
\midrule
Uniform weighting & 85.863\\
Dynamic weighting & \textbf{86.462}\\
\bottomrule
\end{tabular}
\end{table}

Dynamic weighting increases full-test PDMS from 85.863 to 86.462, a gain of 0.600 points (Table~\ref{tab:weight-ablation}). The improvement is measured at the initial-trajectory stage, before expert correction.

\subsubsection{Joint Expert Training}
\label{sec:joint-expert-ablation}
The sequential condition is the training procedure used by \method{}: following preliminary heading training, the position expert is optimized with the heading expert frozen, and the heading expert is then adapted with the position expert frozen. The joint conditions update both experts together during SFT and subsequent GRPO.

\begin{table}[htbp]
\centering\small
\caption{Staged versus joint expert training.}
\label{tab:joint-experts}
\setlength{\tabcolsep}{4pt}
\begin{tabular}{lr}
\toprule
Configuration & PDMS $\uparrow$\\
\midrule
Backbone only & 86.462\\
Staged expert training (\method{}) & \textbf{88.837}\\
Joint SFT & 85.651\\
Joint SFT + GRPO & 83.671\\
\bottomrule
\end{tabular}
\end{table}

The proposed staged training reaches 88.837 PDMS, outperforming joint SFT and joint SFT followed by GRPO by 3.186 and 5.166 points, respectively (Table~\ref{tab:joint-experts}). Both joint-training variants are weaker than the sequential training recipe used by \method{}. These results support the staged optimization of the two experts used in our method.

\subsubsection{Correction-Module Structure and Capacity}
\label{sec:decoder-sharing}
\label{sec:mlp-capacity-ablation}
\paragraph{Parameter-matched MLP expansion.}
To examine whether the gains arise simply from adding parameters, we expand the ordinary MLP trajectory decoder by approximately 156.84M parameters, matching the additional capacity of the two correction experts. The expanded MLP achieves 86.510 PDMS, only 0.048 points above the backbone-only score of 86.462 and 2.327 points below the dual-expert score of 88.837 (Table~\ref{tab:decoder-sharing}). The physically structured correction experts achieve a larger gain than the parameter-matched MLP expansion in this comparison.

\paragraph{Single versus separate experts.}
We replace the two specialized correction experts with a single expert that handles both position and heading correction. The single expert uses one 16-layer autoregressive decoder to generate position residuals and adjacent heading increments, whereas \method{} uses two independent eight-layer experts. Both configurations use the same frozen backbone, 92,614 training scenes, physical output representations, 156.84M correction parameters, and number of training updates.

\begin{table}[htbp]
\centering\small
\caption{Correction-module structure and added capacity on the full test set.}
\label{tab:decoder-sharing}
\resizebox{\columnwidth}{!}{%
\begin{tabular}{lrr}
\toprule
Configuration & Added parameters (M) & PDMS $\uparrow$\\
\midrule
Backbone only & 0 & 86.462\\
MLP decoder expansion & 156.84 & 86.510\\
Separate experts (\method{}) & 156.84 & \textbf{88.837}\\
Single expert, final RL & 156.84 & 84.227\\
Single expert, after SFT & 156.84 & 85.147\\
\bottomrule
\end{tabular}}
\end{table}

Under matched parameter counts and update budgets, \method{}'s separate experts outperform the single expert's final RL endpoint by 4.610 PDMS points (Table~\ref{tab:decoder-sharing}). The single expert also has 7.8\% higher online inference latency. These results favor the complete recipe with separate position and heading experts in the evaluated setting.

\subsubsection{Safety-Gated Progress Reward}
\label{sec:progress-reward-ablation}
We test whether the second position RL stage benefits from reward shaping rather than additional training alone. With a common L1 backbone and heading expert frozen, all variants start from the same first-stage position RL checkpoint and train for five more epochs using the same data, KL reference, optimizer settings, and update budget. Only the coefficient $\lambda$ in Eq.~\eqref{eq:safeep} changes; $\lambda=0$ continues with the original PDMS reward.

\begin{table*}[htbp]
\centering\small
\caption{Safety-gated progress reward ablation on the full test set. Differences are relative to $\lambda=0$.}
\label{tab:progress-reward}
\begin{tabular}{lrr}
\toprule
Reward coefficient $\lambda$ & Test PDMS $\uparrow$ & $\Delta$ PDMS\\
\midrule
0 (original reward) & 88.738 & ---\\
0.5 & 88.749 & $+0.010$\\
1.0 & \textbf{88.840} & $+0.101$\\
2.0 & 88.720 & $-0.018$\\
\bottomrule
\end{tabular}
\end{table*}

The coefficient $\lambda=1$ gives the highest full-test point estimate of 88.840 PDMS, a gain of 0.101 over the original-reward control (Table~\ref{tab:progress-reward}). At $\lambda=2$, the test score is slightly below the original-reward control.

\section{Performance Analysis}
\label{sec:qualitative}
\begin{figure*}[htbp]
\centering
\includegraphics[width=\textwidth]{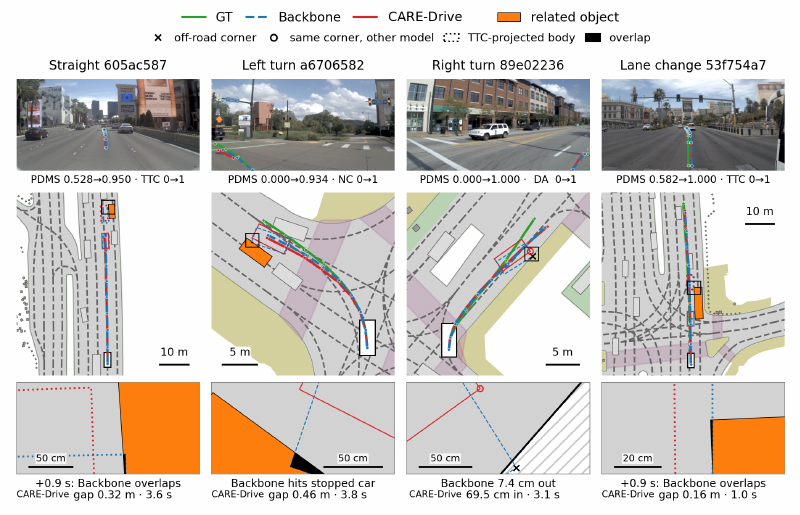}
\caption{\textbf{Qualitative comparison of backbone and complete-system predictions.} Columns show straight driving, a left turn, a right turn, and a lane change. Green denotes the reference trajectory, dashed blue the backbone, and red \method{}. Orange marks related objects; dotted vehicle outlines denote TTC projections, and black regions mark overlaps. Crosses mark off-road corners and circles the corresponding corners of the other prediction. Local enlargements highlight the geometric changes in these selected improvement cases. Scene PDMS is shown on the $[0,1]$ scale.}
\label{fig:qualitative}
\end{figure*}

Figure~\ref{fig:qualitative} compares backbone and \method{} trajectories in straight-driving, left-turn, right-turn, and lane-change scenes. Position and heading corrections change vehicle footprints relative to road boundaries and nearby objects, improving collision, drivable-area, and TTC compliance in the illustrated cases. Additional examples are provided in Appendix~\ref{app:qualitative}.

\begin{figure*}[p]
\centering
\includegraphics[width=\textwidth,height=0.76\textheight,keepaspectratio]{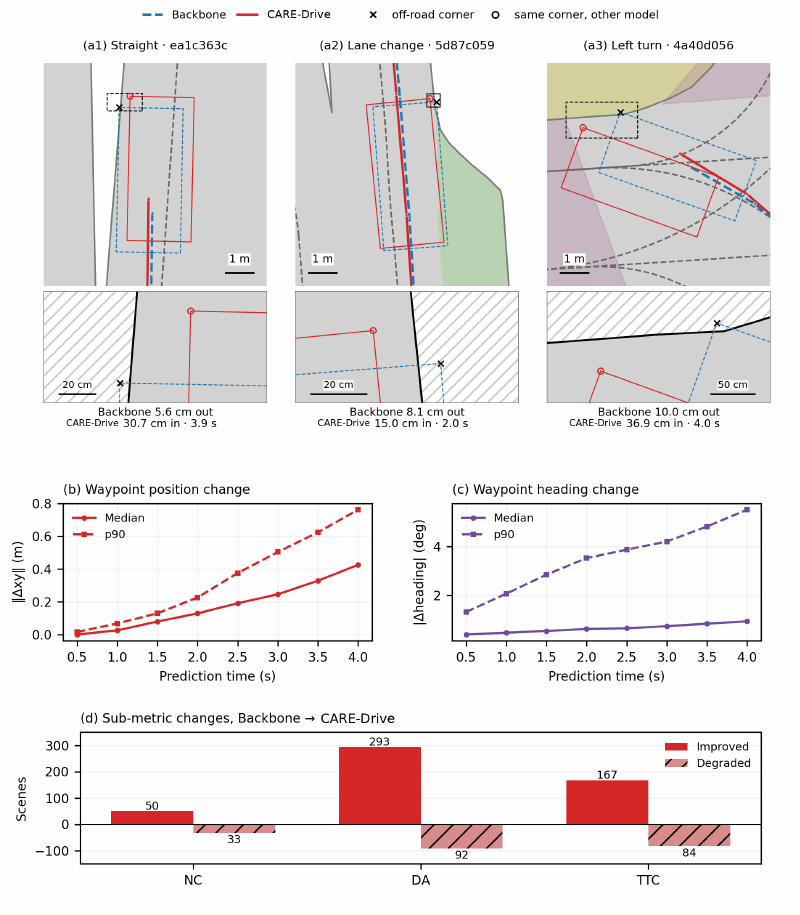}
\caption{\textbf{Geometry, magnitude, and safety outcomes of correction.} (a) Vehicle-footprint examples near road boundaries, with local enlargements. (b,c) Median and 90th-percentile absolute position and heading changes relative to the backbone, by prediction time. (d) Improved and degraded scene counts for collision compliance (NC), drivable-area compliance (DA), and time-to-collision compliance (TTC). The statistics use a random sample from the test set.}
\label{fig:correction-analysis}
\end{figure*}

Figure~\ref{fig:correction-analysis} summarizes correction magnitudes and safety-component changes on a random sample from the test set. The median and 90th-percentile position and heading changes relative to the backbone increase with the prediction horizon. Improved/degraded scene counts are 50/33 for NC, 293/92 for drivable-area compliance, and 167/84 for TTC, with improvements outnumbering degradations in all three components.

\clearpage
\section{Discussion}
\paragraph{Predictive supervision and action interfaces.}
The system demonstrates how future-feature supervision can coexist with a fully observation-based inference path and separate trajectory controls.

\paragraph{Cross-attention readout.}
We also tested an additional cross-attention readout adapter, but it did not yield better planning performance. The final model therefore uses no additional cross-attention readout adapter. Detailed experimental metrics are reported in Appendix~\ref{sec:internal-systems}.

\paragraph{Safety objectives and tail behavior.}
The supplementary reward experiments in Appendix~\ref{app:additional-results} highlight a practical safety tradeoff. L1 error is continuous and averaged over coordinates, whereas NC and DA multiply the planning reward and can change discontinuously with trajectory geometry. In the heading-penalty experiment, scenes with NC below one decrease from 204 to 201, while scenes with DA below one increase from 497 to 522. The two compliance components therefore move in different directions; the corresponding PDMS and component counts are reported together in Appendix~\ref{app:additional-results}.

\section{Conclusion}
We presented \method{}, a future-supervised world--action model combining a deterministic base trajectory with physically meaningful position and heading correction. The complete system achieves 88.837 PDMS on the test set. The ablations support physically structured correction: the proposed staged expert recipe outperforms the tested joint-training, single-expert, and MLP-expansion alternatives. This modular framework provides a basis for extending the approach to more complex, physically meaningful correction experts.

\bibliographystyle{plainnat}
\bibliography{references}

@misc{qwen3vl,
  author={S. Bai and Y. Cai and R. Chen and K. Chen and X. Chen and Z. Cheng and L. Deng and W. Ding and C. Gao and C. Ge and others},
  title={{Qwen3-vl technical report}},
  year={2025},
  howpublished={arXiv preprint arXiv:2511.21631},
  url={https://arxiv.org/abs/2511.21631},
}

@misc{navsim,
  author={D. Dauner and M. Hallgarten and T. Li and X. Weng and Z. Huang and Z. Yang and H. Li and I. Gilitschenski and B. Ivanovic and M. Pavone and others},
  title={{Navsim: data-driven non-reactive autonomous vehicle simulation and benchmarking}},
  year={2024},
  howpublished={Advances in Neural Information Processing Systems 37, pp. 28706–28719},
}

@misc{uniad,
  author={Y. Hu and J. Yang and L. Chen and K. Li and C. Sima and X. Zhu and S. Chai and S. Du and T. Lin and W. Wang and others},
  title={{Planning-oriented autonomous driving}},
  year={2023},
  howpublished={In Proceedings of the IEEE/CVF conference on computer vision and pattern recognition, pp. 17853–17862},
}

@misc{drivelaw,
  author={T. Xia and Y. Li and L. Zhou and J. Yao and K. Xiong and H. Sun and B. Wang and K. Ma and G. Chen and H. Ye and others},
  title={{Drivelaw: unifying planning and video generation in a latent driving world}},
  year={2025},
  howpublished={arXiv preprint arXiv:2512.23421},
  url={https://arxiv.org/abs/2512.23421},
}

@misc{vjepa,
  author={M. Assran and A. Bardes and D. Fan and Q. Garrido and R. Howes and M. Muckley and A. Rizvi and C. Roberts and K. Sinha and A. Zholus and others},
  title={{V-jepa 2: self-supervised video models enable understanding, prediction and planning}},
  year={2025},
  howpublished={arXiv preprint arXiv:2506.09985},
  url={https://arxiv.org/abs/2506.09985},
}

@misc{law,
  author={Y. Li and L. Fan and J. He and Y. Wang and Y. Chen and Z. Zhang and T. Tan},
  title={{Enhancing end-to-end autonomous driving with latent world model}},
  year={2024},
  howpublished={arXiv preprint arXiv:2406.08481},
  url={https://arxiv.org/abs/2406.08481},
}

@misc{wote,
  author={Y. Li and Y. Wang and Y. Liu and J. He and L. Fan and Z. Zhang},
  title={{End-to-end driving with online trajectory evaluation via bev world model}},
  year={2025},
  howpublished={In Proceedings of the IEEE/CVF International Conference on Computer Vision, pp. 27137–27146},
}

@misc{recogdrive,
  author={Y. Li and K. Xiong and X. Guo and F. Li and S. Yan and G. Xu and L. Zhou and L. Chen and H. Sun and B. Wang and others},
  title={{Recogdrive: a reinforced cognitive framework for end-to-end autonomous driving}},
  year={2025},
  howpublished={arXiv preprint arXiv:2506.08052},
  url={https://arxiv.org/abs/2506.08052},
}

@misc{epona,
  author={K. Zhang and Z. Tang and X. Hu and X. Pan and X. Guo and Y. Liu and J. Huang and L. Yuan and Q. Zhang and X. Long and others},
  title={{Epona: autoregressive diffusion world model for autonomous driving}},
  year={2025},
  howpublished={In Proceedings of the IEEE/CVF International Conference on Computer Vision, pp. 27220–27230},
}

@misc{drivevla,
  author={Y. Li and S. Shang and W. Liu and B. Zhan and H. Wang and Y. Wang and Y. Chen and X. Wang and Y. An and C. Tang and others},
  title={{DriveVLA-w0: world models amplify data scaling law in autonomous driving}},
  year={2025},
  howpublished={arXiv preprint arXiv:2510.12796},
  url={https://arxiv.org/abs/2510.12796},
}

@misc{uniworld,
  author={Q. Liu and H. Xu and J. Li and B. Sun and Z. Hao and D. She and X. Zhu and L. Zhang},
  title={{Uni-world vla: interleaved world modeling and planning for autonomous driving}},
  year={2026},
  howpublished={arXiv preprint arXiv:2603.27287},
  url={https://arxiv.org/abs/2603.27287},
}

@misc{hydra,
  author={Z. Li and K. Li and S. Wang and S. Lan and Z. Yu and Y. Ji and Z. Li and Z. Zhu and J. Kautz and Z. Wu and others},
  title={{Hydra-mdp: end-to-end multimodal planning with multi-target hydra-distillation}},
  year={2024},
  howpublished={arXiv preprint arXiv:2406.06978},
  url={https://arxiv.org/abs/2406.06978},
}

@misc{diffusiondrive,
  author={B. Liao and S. Chen and H. Yin and B. Jiang and C. Wang and S. Yan and X. Zhang and X. Li and Y. Zhang and Q. Zhang and others},
  title={{Diffusiondrive: truncated diffusion model for end-to-end autonomous driving}},
  year={2025},
  howpublished={In Proceedings of the Computer Vision and Pattern Recognition Conference, pp. 12037–12047},
}

@misc{pwm,
  author={Z. Zhao and T. Fu and Y. Wang and L. Wang and H. Lu},
  title={{From forecasting to planning: policy world model for collaborative state-action prediction}},
  year={2025},
  howpublished={arXiv preprint arXiv:2510.19654},
  url={https://arxiv.org/abs/2510.19654},
}

@misc{resworld,
  author={J. Zhang and Z. Fu and Z. Xu and W. Dai and Q. Liu and Y. Wang},
  title={{ResWorld: temporal residual world model for end-to-end autonomous driving}},
  year={2026},
  howpublished={arXiv preprint arXiv:2602.10884},
  url={https://arxiv.org/abs/2602.10884},
}

@misc{grpo,
  author={Z. Shao and P. Wang and Q. Zhu and R. Xu and J. Song and X. Bi and H. Zhang and M. Zhang and Y. Li and Y. Wu and others},
  title={{Deepseekmath: pushing the limits of mathematical reasoning in open language models}},
  year={2024},
  howpublished={arXiv preprint arXiv:2402.03300},
  url={https://arxiv.org/abs/2402.03300},
}

@article{coworld,
 author={Minqing Huang and Yujiao Xiang and Zihan Liang and Jiajie Huang and Jingqi Wang and Yuheng Zhou and Zhi Xu and Feiyang Tan and Hangning Zhou and Mu Yang and Gong Chen},
 title={{CoWorld-VLA: Thinking in a Multi-Expert World Model for Autonomous Driving}},
 journal={arXiv preprint arXiv:2605.10426}, year={2026}, url={https://arxiv.org/abs/2605.10426}
}

@article{autovla,
 author={Zewei Zhou and Tianhui Cai and Seth Z. Zhao and Yun Zhang and Zhiyu Huang and Bolei Zhou and Jiaqi Ma},
 title={{AutoVLA: A Vision-Language-Action Model for End-to-End Autonomous Driving with Adaptive Reasoning and Reinforcement Fine-Tuning}},
 journal={Advances in Neural Information Processing Systems}, year={2025}, url={https://arxiv.org/abs/2506.13757}
}

@article{ppo,
 author={John Schulman and Filip Wolski and Prafulla Dhariwal and Alec Radford and Oleg Klimov},
 title={{Proximal Policy Optimization Algorithms}}, journal={arXiv preprint arXiv:1707.06347}, year={2017}, url={https://arxiv.org/abs/1707.06347}
}

@inproceedings{scst,
 author={Steven J. Rennie and Etienne Marcheret and Youssef Mroueh and Jerret Ross and Vaibhava Goel},
 title={{Self-Critical Sequence Training for Image Captioning}},
 booktitle={Proceedings of the IEEE Conference on Computer Vision and Pattern Recognition}, year={2017},
 url={https://openaccess.thecvf.com/content_cvpr_2017/html/Rennie_Self-Critical_Sequence_Training_CVPR_2017_paper.html}
}

\appendix
\section{Implementation and Training Details}
\label{app:interfaces}
\paragraph{Backbone.}
The visual input uses three separate four-frame videos at $1024\times512$ per frame, with the recorded frame indices and FPS. The visual tower, merger, and associated visual projections are frozen; the language backbone, token embeddings, future projection, and trajectory MLP are trained. The forward pass uses BF16 while trainable parameters and AdamW states use FP32. The 96 future query positions use four temporal token identities repeated 24 times each. The eight action queries have distinct token identities. Backbone optimization combines trajectory regression and future-feature supervision.

Backbone AdamW uses $\beta_1=0.9$, $\beta_2=0.95$, $\epsilon=10^{-8}$, weight decay 0.01, and gradient clipping at one. Peak learning rates are $2\times10^{-5}$ for the language module, $10^{-4}$ for the future projection, and $10^{-5}$ for the trajectory head. A continuous cosine schedule follows 5\% warmup. The effective batch is 128, with microbatch one, eight accumulation passes, and 16 devices. The five epochs contain 725 updates each, including a final batch of 100 real scenes. Dynamic weights are applied before averaging over the real scenes in the global batch.

\paragraph{Expert memories.}
The position expert projects eight action features and eight base-trajectory features into separate memory token types. Base features concatenate $x/50$, $y/20$, adjacent displacement divided by ten, $\sin\psi$, $\cos\psi-1$, and normalized time $t/8$. Memory type and temporal embeddings are added before normalization. The autoregressive decoder includes causal self-attention and cross-attention to this memory. Position tokens also use time and axis embeddings.

The heading expert uses the same action features and corrected-XY features: scaled XY, scaled adjacent displacement, and $t/8$. It has separate memory type and position embeddings. Heading labels are formed by wrapping adjacent ground-truth angle differences, clipping to $\pm20^\circ$, and rounding to 0.1-degree units. Decoding sums increments before angle wrapping.

\paragraph{Quantization range analysis.}
For position residuals, we analyzed the target-minus-backbone XY differences in 92,614 training scenes from the preceding L1-backbone configuration. With $b_x=5$ m, $b_y=3$ m, and 401 uniformly spaced values per axis, clipping affected 0.4157\% of scenes. Reconstructing the target trajectories from quantized residuals gave an ADE of 0.0088 m. Increasing $b_x$ to 10 m with the same number of values reduced the clipped-scene fraction to 0.1890\% and increased the reconstruction ADE to 0.0138 m. This comparison characterizes the coverage--resolution tradeoff underlying the adopted position bounds.

For heading increments, we analyzed 500 uniformly sampled training scenes, containing 4,000 adjacent ground-truth heading differences at 0.5-second intervals. Differences were angle-wrapped before quantization. All 4,000 increments lay within $[-20^{\circ},20^{\circ}]$. Quantizing them in $0.1^{\circ}$ steps and accumulating the increments reconstructed the ground-truth heading sequences with an MAE of $0.0468^{\circ}$. This analysis provides training-sample coverage and quantization-error evidence for the adopted heading range.

The codec uses 401 uniformly spaced values including both endpoints, giving 400 intervals. A symmetric bound $b$ therefore yields a step of $2b/400$: 0.025 m for X residuals, 0.015 m for Y residuals, and $0.1^{\circ}$ for heading increments.

\paragraph{Reinforcement stages.}
Expert RL uses FP32, effective batch 128 (microbatch eight on each of 16 devices), and eight stochastic candidates per scene. Each epoch has 724 updates, with 70 real scenes in the final batch. A five-epoch stage therefore has 3,620 updates, 463,070 scene visits, and 3,704,560 stochastic candidate evaluations. The scene visits accumulate over five passes through the training set. Final heading adaptation additionally generates one detached greedy-baseline trajectory per scene visit.

Expert RL uses AdamW with learning rate $10^{-5}$, $\beta=(0.9,0.999)$, $\epsilon=10^{-8}$, zero weight decay, constant learning rate, and gradient clipping at one. The optimizer and stage random stream are reset at stage boundaries; each stage retains its fixed policy reference. The backbone seed is 42 and the expert seed is 20261019. All stages use the listed fixed seeds.

\begin{table*}[t]
\centering\small
\caption{Training lineage of the dynamic system. Every row is a five-epoch stage. Each stage retains its final-epoch checkpoint; the table reports the corresponding validation scores. R0 later repeats C3 and D1 from the recorded dynamic parents.}
\label{tab:curriculum}
\footnotesize\setlength{\tabcolsep}{3.5pt}
\begin{tabular}{llllrr}
\toprule
Stage & Updated component & Objective / role & Fixed context & Updates & Val497 PDMS\\
\midrule
A0 & Backbone & Dynamic weighted L1 + future features & Frozen visual tower & 3,625 & 87.611\\
B1 & Heading & Token SFT & Backbone; base XY & 3,620 & 87.527\\
B2 & Heading & First official-reward RL stage & Backbone; base XY & 3,620 & 89.581\\
B3 & Heading & Second official-reward RL stage & Backbone; base XY & 3,620 & 90.126\\
C1 & Position & Residual-token SFT & Backbone; B3 heading helper & 3,620 & 90.128\\
C2 & Position & Official reward; global RMS & Backbone; B3 heading helper & 3,620 & 90.924\\
C3 & Position & Safety-gated progress; $\lambda=1$ & Backbone; B3 heading helper & 3,620 & 91.043\\
D1 & Heading & Greedy baseline; RMS matching & Backbone; final C3 XY & 3,620 & 90.675\\
\bottomrule
\end{tabular}
\end{table*}

\section{Historical Selection and Statistical Interpretation}
\label{app:selection}
The historical expert system uses the fourth-epoch L1 backbone, a position endpoint internally named P11, and a heading endpoint named P18. Position-parent selection and the P14--P25 heading exploration consulted the small-test-set results from strategy ablations. The later dynamic and cross-attention branches fix their final-epoch weights before the full-split expansion, but inherit those recipe choices. R0 repeats the endpoint training with the same seed.

The full-split historical-minus-L1 interval and the dynamic-minus-historical interval answer different questions. The former concerns a fixed complete expert chain relative to its associated base policy. The latter concerns a new complete training recipe relative to the historical expert system. The cross-attention branch likewise compares complete systems with their respective training histories.

The reported dynamic-versus-historical interval is $[-0.336,0.410]$, taken from the accepted three-system report. A separate accepted R0 report gives $[-0.342,0.397]$ for the same point difference under its own bootstrap run. Both intervals span zero. The intervals describe log-level resampling variability for fixed trained checkpoints.

For R1 and R2, the unadjusted intervals are $[-0.354,0.052]$ and $[-0.354,-0.012]$. The two prespecified comparisons use Bonferroni-adjusted 97.5\% intervals of $[-0.383,0.078]$ and $[-0.379,0.014]$, respectively.

\section{Additional Results}
\label{app:additional-results}
This appendix collects historical system comparisons and exploratory safety objectives that supplement the main ablations.

\subsection{Historical System Comparison}
\label{sec:internal-systems}
\begin{table}[t]
\centering\small
\caption{Internal system comparison on the test set. H denotes the historical expert system.}
\label{tab:systems}
\begin{tabular}{lr}
\toprule
System & Test PDMS\\
\midrule
L1 backbone, epoch 4 & 85.988\\
Historical experts (H) & 88.810\\
Cross-attention & 88.726\\
Dynamic / R0 & \textbf{88.837}\\
\bottomrule
\end{tabular}
\end{table}

\begin{figure*}[t]
\centering\includegraphics[width=\textwidth]{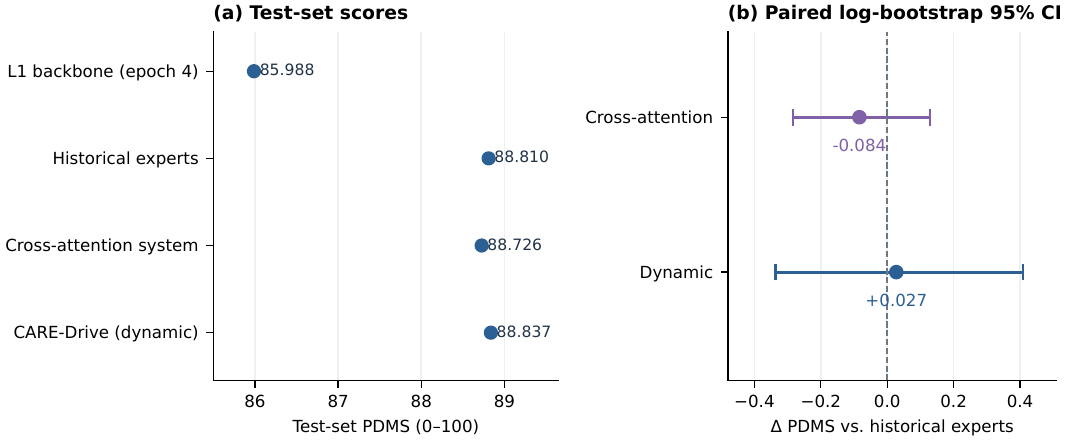}
\caption{\textbf{Test-set performance of complete systems.} (a) PDMS of each fixed system. (b) Paired log-bootstrap 95\% intervals for differences from the historical expert system. Intervals are computed by resampling logs for fixed trained models.}
\label{fig:full}
\end{figure*}

Table~\ref{tab:systems} and Figure~\ref{fig:full} summarize the historical system comparisons. The historical expert system adds 2.821 PDMS over its corresponding L1 backbone on the full split, with a paired log-bootstrap 95\% interval of $[2.165,3.452]$. This comparison measures the performance gain of the complete expert chain.

The dynamic system reaches 88.837, the highest full-split point estimate among these internal systems. Its difference from the historical expert system is only $+0.027$, with a 95\% interval of $[-0.336,0.410]$. The cross-attention system differs by $-0.084$, with an interval of $[-0.284,0.128]$. Both intervals cross zero. These branches use complete-system retraining with their respective initializations and training histories. The repeated R0 endpoint reproduces the dynamic score.

\subsection{Controlled Safety-Penalty Interventions}
We freeze the dynamic backbone and the relevant preliminary expert parents, and train an actual shared control R0. R1 modifies the final position reward by subtracting
\begin{equation}
c_i=(1-\mathrm{NC}_i)+(1-\mathrm{DA}_i)
\end{equation}
with coefficient one, then applies the original heading-adaptation recipe to its own positions. R2 shares R0's position policy and fixed XY, and subtracts the same cost from both sampled and greedy-baseline rewards in the heading stage. Other settings, including the official reference advantage RMS, remain fixed. R0 and R1 each train two stages; R2 trains one new heading stage, with the shared R0 position stage counted once. Together, these interventions add five stages and 18,100 updates.

\begin{table}[t]
\centering\small
\caption{Controlled risk experiments. All use the same dynamic backbone. Intervals are Bonferroni-adjusted 97.5\% paired log-bootstrap intervals for full-split differences from R0.}
\label{tab:risk}
\begin{tabular}{lrrr}
\toprule
Variant & Val497 & Full PDMS & $\Delta$ vs. R0\\
\midrule
R0: original recipe & 90.675 & \textbf{88.837} & ---\\
R1: position penalty & 90.819 & 88.682 & $-0.155$\\
R2: heading penalty & 90.949 & 88.655 & $-0.182$\\
\midrule
\multicolumn{4}{l}{R1 interval: $[-0.383,\ 0.078]$}\\
\multicolumn{4}{l}{R2 interval: $[-0.379,\ 0.014]$}\\
\bottomrule
\end{tabular}
\end{table}

\begin{figure*}[t]
\centering\includegraphics[width=\textwidth]{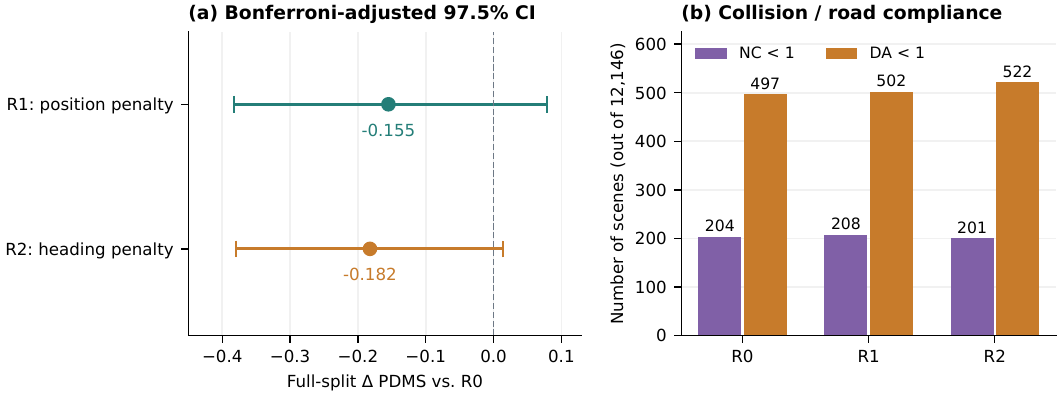}
\caption{\textbf{Planning scores and compliance under safety reward penalties.} (a) Both full-split differences are negative in point estimate, with two-comparison-adjusted intervals crossing zero. (b) A small reduction in collision-compliance failures in R2 accompanies more road-compliance failures. Counts indicate components below one, including partial NC penalties.}
\label{fig:risk}
\end{figure*}

Neither penalty variant improves full-split PDMS or mean risk cost (Table~\ref{tab:risk}). R0, R1, and R2 have average costs 0.0567, 0.0575, and 0.0585, respectively. R2 reduces the number of scenes with NC below one from 204 to 201, but increases the number with DA below one from 497 to 522. Relative to R0, R1 and R2 introduce 75 and 61 new scenes whose NC--DA product becomes zero. Figure~\ref{fig:risk} reports both adjusted uncertainty and component counts.

The unadjusted 95\% interval for R2 excludes zero, but the prespecified two-comparison correction yields an interval crossing zero. Both interventions yield lower observed mean PDMS than R0, with adjusted intervals spanning zero.

\subsection{Training-Time Safety Weighting and Conditional Stopping}
\label{sec:ilc}
The original dynamic rule requires extra frozen-endpoint inference. We additionally test IL-C, which records detached training-forward predictions during epochs one through four and uses their NC/DA outcomes to fix the next epoch's weights. A scene is difficult when predicted NC is lower than GT NC or predicted DA is lower than GT DA. Difficult and other scoreable scenes receive raw weights two and one, followed by mean-one normalization; the 158 neutral scenes remain at weight one. IL-C reuses training-forward predictions for scoring. These predictions come from the successive model states encountered during optimization.

Before training, continuation to seven new expert stages was conditioned on both a validation PDMS of at least 87.611 and a strictly lower mean safety cost than the dynamic backbone. The new backbone completes five epochs but scores 87.534, a decline of 0.076 points. Both policies have an exact discrete cost sum of 30 over 497 validation scenes. Both average costs equal $30/497$.

Under the prespecified continuation criteria, IL-C ended after the five-epoch backbone stage; its evaluation therefore covers this backbone endpoint.

\subsection{Additional Safety Statistics}
\begin{table}[h]
\centering\small
\caption{Safety costs and component counts on the full split. NC below one includes partial penalties.}
\begin{tabular}{lrrr}
\toprule
Variant & NC $<1$ & DA $<1$ & Mean cost\\
\midrule
R0 & 204 & 497 & 0.0567\\
R1 & 208 & 502 & 0.0575\\
R2 & 201 & 522 & 0.0585\\
\bottomrule
\end{tabular}
\end{table}

The risk experiments evaluate complete systems, while IL-C evaluates the backbone endpoint described in Section~\ref{sec:ilc}.

The two safety-focused research batches comprise five correction stages plus one IL-C backbone stage, or 21,725 updates.

\clearpage
\onecolumn
\section{Additional Qualitative Examples}
\label{app:qualitative}
\begin{figure}[htbp]
\centering
\includegraphics[width=\textwidth,height=0.76\textheight,keepaspectratio]{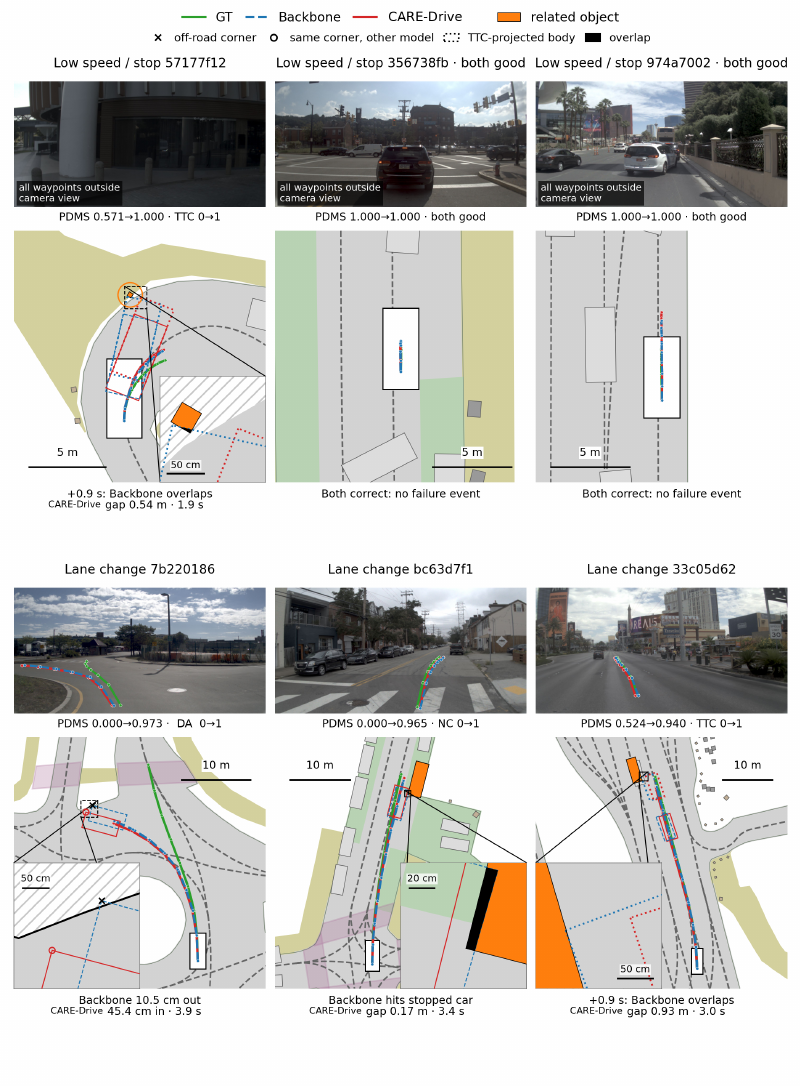}
\caption{\textbf{Additional low-speed and lane-change examples.} The upper row includes one improvement case and two cases with unchanged maximum PDMS; the lower row shows lane-change improvements. For examples labeled ``All waypoints outside camera view'', the trajectory positions are shown in the bird's-eye-view panels.}
\label{fig:more-examples-1}
\end{figure}

\begin{figure}[htbp]
\centering
\includegraphics[width=\textwidth,height=0.76\textheight,keepaspectratio]{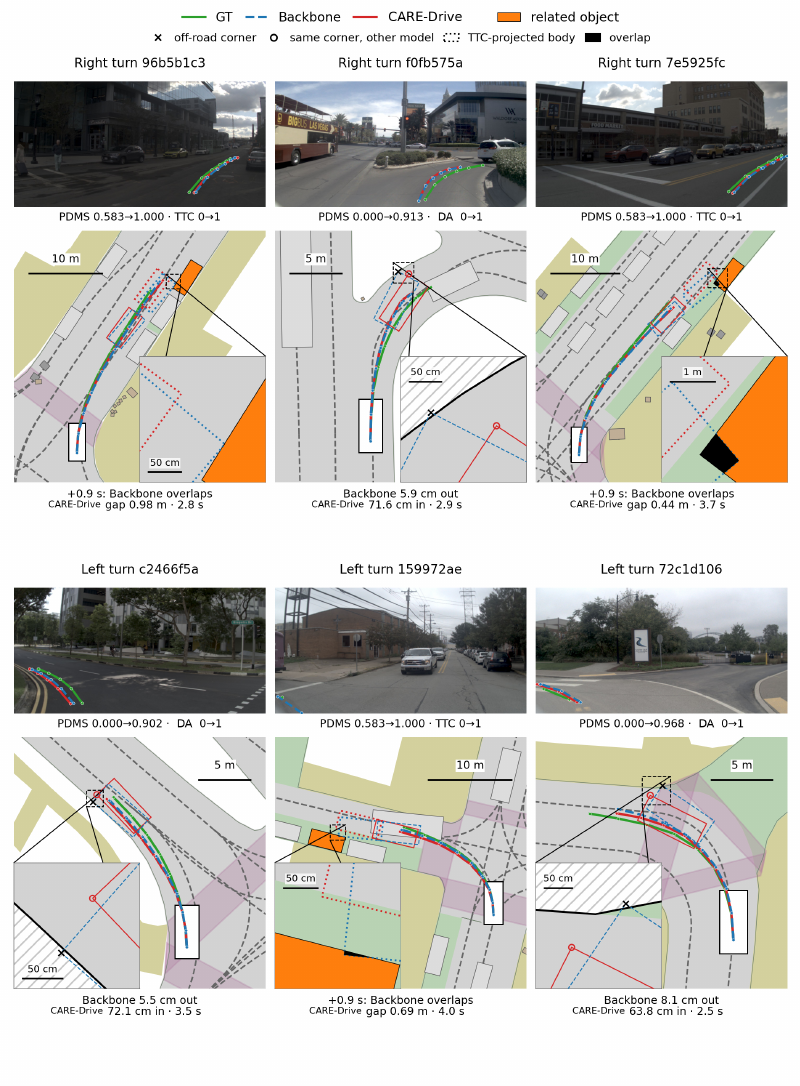}
\caption{\textbf{Additional turning examples.} Right turns appear in the upper row and left turns in the lower row. Local enlargements highlight road-boundary and TTC-projection differences associated with the annotated improvements.}
\label{fig:more-examples-2}
\end{figure}

\begin{figure}[htbp]
\centering
\includegraphics[width=\textwidth,height=0.76\textheight,keepaspectratio]{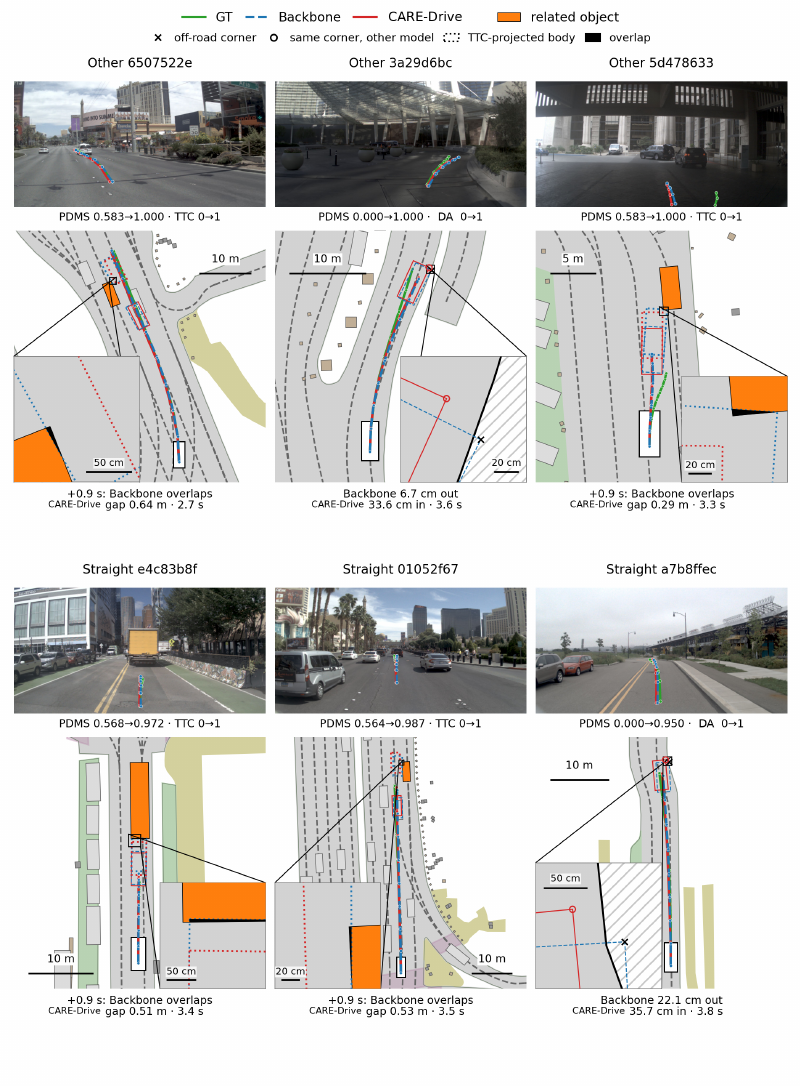}
\caption{\textbf{Additional miscellaneous and straight-driving examples.} The upper row retains the supplied ``Other'' category; the lower row shows straight-driving cases.}
\label{fig:more-examples-3}
\end{figure}

Figures~\ref{fig:more-examples-1}--\ref{fig:more-examples-3} extend the selected examples in Section~\ref{sec:qualitative}. They retain the same colors, footprint markers, and scene-level $[0,1]$ PDMS scale. The examples cover low-speed or stopped traffic, lane changes, turns, straight driving, and scenes labeled ``Other'' in the supplied grouping. The two ``both good'' examples illustrate unchanged maximum scores.

\clearpage

\clearpage

\end{document}